\documentclass[11pt]{article}
\usepackage{graphicx} 
\usepackage{fullpage} 
\usepackage{setspace} 
\usepackage[bottom]{footmisc} 
\usepackage[comma,authoryear,round,sort]{natbib} 
\usepackage{url} 
\usepackage{titlesec}

\titleformat{\subsection}{\itshape}{\thesubsection}{1em}{\large}[]

\title{The Synthetic Imputation Approach: Generating \\ Optimal Synthetic Texts For Underrepresented  \\ Categories In Supervised Classification Tasks \vspace{0.5cm}}
\author{Joan C. Timoneda\thanks{Assistant Professor of Political Science, Purdue University. Contact: timoneda@purdue.edu.}}
\date{}

\begin{document}

\maketitle

\begin{abstract}
\noindent Encoder-decoder Large Language Models (LLMs), such as BERT and RoBERTa, require that \textit{all} categories in an annotation task be sufficiently represented in the training data for optimal performance. However, it is often difficult to find sufficient examples for all categories in a task when building a high-quality training set. In this article, I describe this problem and propose a solution, the synthetic imputation approach. Leveraging a generative LLM (GPT-4o), this approach generates synthetic texts based on careful prompting and five original examples drawn randomly with replacement from the sample. This approach ensures that new synthetic texts are sufficiently different from the original texts to reduce overfitting, but retain the underlying substantive meaning of the examples to maximize out-of-sample performance. With 75 original examples or more, synthetic imputation's performance is on par with a full sample of original texts, and overfitting remains low, predictable and correctable with 50 original samples. The synthetic imputation approach provides a novel role for generative LLMs in research and allows applied researchers to balance their datasets for best performance.

\end{abstract}

\thispagestyle{empty}

\pagebreak

\setcounter{page}{1}

\setstretch{2}


\noindent One of the key requirements to fine-tune supervised deep learning Transformers models like BERT and RoBERTa is high-quality training data. Yet many applied datasets suffer from imbalance problems: while researchers often have sufficient text instances for one or more categories, they do not for others. In fact, it is generally challenging to find sufficient representative examples for each category in the training set. In an annotation task with multiple categories, finding sufficient examples for each category may not be feasible due to resource constraints. Also, some categories may be rare events, making it inherently difficult to extract sufficient examples for a training set. The problem is that training a model with underrepresented categories in unbalanced datasets often results in low levels of performance for those categories and for the overall model.

This article proposes a \textit{synthetic imputation} solution for this particular problem. The process leverages a generative Large Language Model (LLM), such as OpenAI's GPT-4o, to produce synthetic texts for underrepresented categories in an annotation task, helping to populate these categories with sufficient examples to reach optimal levels of performance. In doing so, the approach resembles the philosophy of multiple imputation, where information from a smaller number of examples can inform the imputation of a larger sample. It also draws from the data augmentation literature in computer science, where algorithms have been recently developed to improve synthetic text generation \citep{ng2020ssmba, wei2021text}. However, these novel approaches often produce text that is too similar in nature and structure to the examples provided. Highly analogous texts can lead to overestimating true model performance, as the model classifies them more accurately due to textual resemblance, not its ability to understand the underlying textual dimensions within a particular category. This compromises the model's capacity to generalize to unseen data.\footnote{In this article, I will refer to this phenomenon as `overfitting' because it resembles overfitting in general machine learning applications: the model learns too well on the training data and then fails to generalize well to unseen data. Overly similar training textual  examples, therefore, lead to a form of overfitting.} Therefore, the key for the synthetic imputation approach to produce reliable out-of-sample performance is that synthetic texts be both (1) sufficiently similar to improve classifier performance and (2) sufficiently different from each other to avoid repetition and overfitting. To achieve this, synthetic imputation uses five random examples from the original sample together with a carefully crafted prompt to generate each synthetic text.\footnote{In this article, I use OpenAI's GPT-4o model. Models such as Llama or Claude are also good alternatives (citation omitted).} The prompt includes specific instructions to write different text in terms of sentence structure and key content, while preserving the overall common ideas across the five examples. To increase variability in LLM responses, the five examples are drawn anew and with replacement for each synthetic text the LLM produces. With this prompt design, each new synthetic text is unique and different from the original samples, but retains the substantive ideas underlying the different examples. To understand how the synthetic imputation approach compares to current alternatives in the literature, I provide performance metrics for a current state-of-the-art data augmentation approach: Self-supervised Manifold Based Augmentation, or SSMBA \citep{ng2020ssmba}.

The results show that the synthetic imputation approach significantly outperforms SSMBA and produces performance scores similar those of a larger dataset with more optimal numbers of original observations. First, when synthetic imputation can draw from at least 75 original observations, its category-specific performance is on par with datasets with 151 and 218 original observations (the datasets tested in this article). Generating 76 and 143 synthetic observations, respectively, is equivalent to having the full set of original texts. Importantly, even with only 50 original observations, synthetic imputation's overfitting levels remain low ---between 2 and 4\%--- and always in a positive direction, allowing researchers to report a predictable penalized performance metric.\footnote{This will still produce slightly biased out-of-sample annotations, but the amount of measurement bias will be relatively small and preferable over the large amounts of bias induced by low observations.} Low overfitting levels with 50 original observations also translate into much lower measurement bias in annotation tasks when compared to any of the two alternatives proposed in this article. First, with no synthetic data, a category with only 50 original observations will underperform significantly --by 74.5\% and 59.5\% in our testing. Second, while SSMBA yields good performance, it tends to generate texts similar to one another in terms of sentence structure and content, leading to overfitting. In fact, when to compared to SSMBA, synthetic imputation reduces overfitting by 6 to 15\% on average in our testing, achieving scores much closer to true out-of-sample performance.

This article makes three major contributions. First, it proposes a solution in terms of data augmentation within political science, a field where text analysis using machine and deep learning is becoming more common \citep{hobbs2019effects, roberts2016navigating, egami2023using, rodriguez2022word, rodriguez2023embedding, grimmer2021machine, chang2020using, grimmer_stewart_2013, barbera2021automated, catalinac2016pork, rice2021corpus, timoneda_jop, timoneda2025behind}. By proposing an easy-to-implement, intuitive method to generate synthetic training data without sacrificing performance, this article addresses a major need in current research in the discipline using LLMs. For many applied researchers, this novel approach means that rather than abandoning projects or resorting to inefficient groupings of categories, they now can augment their data with little or no overfitting concerns using accessible generative LLMs and combining them with a BERT or RoBERTa classifier.\footnote{Researchers could also use one generative LLM to generate synthetic data, and a separate generative LLM to produce the annotations.} Second, this article expands upon computer scientists' efforts by proposing a data augmentation approach that is simpler to implement than current alternatives and improves out-of-sample performance in annotation tasks. Third, the synthetic imputation approach proposes a novel role for generative LLMs in data augmentation, tapping into their significant potential to generate new text, and is the first one ---in the author's knowledge--- to do so.

\section*{The Problem}

Large language models (LLMs) used for supervised text classification such as BERT and RoBERTa require a critical number of observations per category to perform well \citep{timoneda_jop}. The critical number is task-dependent, but there should generally be sufficient instances for each category to be represented multiple times in the trainer batches, which are small subsets of data that are passed through the neural network during training.\footnote{Model training for BERT and RoBERTa is done in batches. Batches are splits of the data that are fed into the neural network. A training set with 1,000 observations and a batch size of 8 will have 125 batches of 8 observations. With a batch size of 16, the total number of batches would be 63, 62 with 16 observations and 1 batch with 8.} The most common batch sizes are 16 or 32 observations,\footnote{Batch sizes are usually selected in powers of 2. Performance does not vary significantly when using 16 or 32 batch sizes \citep{timoneda_jop}.} and each batch should contain at least a few observations of all categories \citep{liu2019roberta}.  With smaller batch sizes or with a larger number of categories in the data, it may often be the case that some categories are not represented in all the batches if data are scarce \citep{zhou2021isobn}.\footnote{For instance, if a category only has 40 training examples and the batch size is set to 16, its per-batch average will be 2.5 and some batches will have zero observations for that category.} Generally, more examples tend to improve model performance, following a diminishing returns curve.  

Obtaining sufficient high-quality samples \textit{per category} can be difficult, especially if the number of categories is relatively large. A research team may not have the resources to produce a high-N balanced training set, or some categories may be rare events. By way of an example, say we need to classify a set of texts of medium complexity into 10 categories from an initial corpus of 10 million texts. We could manually code around 2,000 observations with as much balance as possible across the 10 categories. During fine-tuning, with a batch size of 16,\footnote{The batch size should be greater than the number of categories, so a batch size of 8 would not be recommended if there are 10 categories.} 200 observations per category will result in samples for each category appearing 1.6 times per batch on average (125 batches in total), with some batches having 2, 3 or 4 observations and some having none. With a batch size of 32, each category will appear 3.2 times on average. In both cases, the gradient will be calculated with sufficient information to converge toward the global minimum and the model should perform reasonably well.\footnote{More data is likely to result in better performance, but BERT and RoBERTa models perform best with batch sizes of 16 and 32 (48 in some tasks) \citep{liu2019roberta}. See \citet{timoneda_jop} for more detail on model performance with smaller samples.}  If one category only has 50 examples, however, examples of this category will only appear in a limited number of batches (50 at most). When they do appear, the batches most often will only have 1 example of that category, and many batches will have none. This will reduce overall model performance and significantly decrease F1 scores for the category, as the overall gradients are calculated without taking into account sufficient information for the underrepresented category.

These issues have been the focus of intense debate in computer science in recent years, where researchers have developed various approaches in the relatively nascent field of data augmentation for text data. One approach have been rules-based algorithms, where tokens are swapped, inserted and deleted at random to generate a similar but different text sample from one original text \citep{wei2019eda, wei2021text}. A second approach has been example interpolation, where tokens from two or more samples are mixed to generate a new synthetic example \citep{guo2020nonlinear}. A third and more complex approach uses deep learning models to generate augmented data. \citet{nie2020named} use pretrained embeddings to generate augmented data based on semantic neighbors in a sentence, while \citet{ng2020ssmba} use autoencoders to deconstruct and reconstruct sentences using BERT \citep{devlin2018bert}. However, these models were developed before the rise of generative AI models, which in essence replicate many of the augmentation behaviors these authors identify. They also produce texts that remain similar in both grammar and vocabulary to the original texts, failing to introduce sufficient variation in the new texts, which can negatively affect out-of-sample performance. Therefore, I contend that a simpler, more intuitive approach to synthetic data augmentation is to use a generative model to create substantively similar but grammatically and semantically different text samples that maximize performance and reduce overfitting.

\section*{A Solution: The Synthetic Imputation Approach}

The logic for the synthetic imputation approach is similar to multiple imputation: if sufficiently informative data exist, one should be able to generate imputations that approximate the true but unknown value of missing observations. The advent of generative LLMs makes it possible to impute training data in supervised text classification. In this article, I use GPT-4o because its performance is strong and it is widely used in the social sciences, but models such as Llama or Claude are good alternatives (citation omitted).\footnote{Meta's Llama 3 family of generative models is often preferred by some researchers due to their open-source nature and free access.}

The synthetic imputation approach has four steps. First is to find \textit{at least} 50 text instances for each category, especially those which are less frequent in the data.\footnote{Below 50 observations, the model is likely to increasingly overfit the data as there are not sufficiently different texts to produce varied synthetic responses, leading to very similar texts that the model can easily identify. This inflates performance metrics but reduces the model's ability to generalize to a larger, unseen sample.} Second, a generative model (in our case, GPT-4o) produces a set of texts that resemble the original set of examples, using a general prompt and five examples drawn at random from the sample \textit{for each generated text.} Ideally, the sum of the original and synthetically imputed texts will be at least 200 to ensure sufficient category representation in batches of 16 or 32.\footnote{Performance is still likely to increase above 200 observations, but with clear diminishing returns. The reason is that more representation in the batches will produce better performance, up to a point, but with 200 observations and a batch size of 16 or 32, RoBERTa models have been shown to produce good overall performance \citep{timoneda_jop}. Still, this article leaves the door open for the researcher to decide the amount of data required for each specific task to maximize performance through testing. Some tasks may require more or less observations, and the researcher can run different cross-validation trials on similar text data to approximate model performance with various data sizes.} Third, the researcher supervises the text generation process. They ensure that the texts are (1) sufficiently similar to ensure the training data is high-quality, and (2) sufficiently different to avoid repetition and, consequently, overfitting. Validation of the procedure may lead to modifications in the prompt and the examples to improve overall model performance. Indeed, overfitting is a major concern, as texts with high levels of similarity will make it easier for the classifier to identify them in the training and validation data, resulting in inflated performance metrics and unreliable out-of-sample generalizations. Accurate prompting and clear random examples should aid in introducing greater amounts of variation in synthetic text output without compromising the substantive meaning of the synthetic texts. Fourth, the team fine-tunes a final BERT or RoBERTa model to their specific task. Please see Figure \ref{fig:chart} (page 11) for a flow chart describing how to generate a synthetic text using the synthetic imputation approach, with an example drawn from the first dataset used in the analysis. 

In this article, I will contrast the performance of the synthetic imputation approach with one recent state-of-the-art data augmentation algorithm by \citet{ng2020ssmba}: Self-supervised Manifold Based Augmentation (SSMBA). This approach is well-suited to specialized texts for its combined use of neural networks, both autoencoders and Transformers, which aids in contextual understanding and sentence construction. The intuition behind the approach is to use de-noising autoencoders (a type of neural network), which reduce the sentence down to its core meaning to create masked tokens.\footnote{This resembles masked language modeling (MLM), the process by which BERT and RoBERTa models learn.} It then uses a reconstruction function powered by a BERT Transformer model to fill the masked tokens with new, different tokens. A simple example from the authors is as follows. The sentence `I burst through the cabin doors' becomes encoded into `$<$mask$>$ burst through the $<$mask$>$ doors', and then reconstructed by BERT into `He burst through the double doors'. Researchers can increase or decrease the frequency of the masked works, with a default value of 15\% that can be modified up to 80\%. Following \citet{wettig2022should} and \citet{timoneda2025behind}, we use a reconstruction rate of 40\% of tokens for both applications described in the next section.\footnote{Again, this resembles MLM in BERT-like models, where 15\% of tokens were originally masked. Further research has shown that masking 40\% of tokens, both generally or selectively, improves performance \citep{timoneda2025behind, wettig2022should}.} The code to apply SSMBA is also available in the replication materials.

\subsection*{Two Applications: Political Nostalgia and International Speeches}

I illustrate the synthetic imputation procedure with two annotation applications or tasks, and compare all results to SSMBA. The first task uses data on political nostalgia from \citet{muller2024nostalgia}. The authors collected sentences from different party manifestos in 24 European countries. Four human annotators coded the data into either nostalgic or not nostalgic, and the authors aggregated these into a binary variable where a sentence is classified as nostalgic if at least three coders agree. The final dataset contains 1,200 annotated sentences, with 151 nostalgic and 1,049 not nostalgic texts. Sentences from the sample were 20.83 words long on average, with a minimum of 3 and a maximum of 174 words. This dataset represents a binary application with relatively short texts, which is a text structure common in political science applications that use sentence-level and Twitter data.

The second application uses data from the Global Populism Database (GPD) by \citet{hawkins2019measuring}.\footnote{The dataset and materials are available at: \url{https://dataverse.harvard.edu/dataset.xhtml?
persistentId=doi:10.7910/DVN/LFTQEZ}} Started in 2006, the project created a large dataset of populist speeches by world leaders. The dataset contains 843 speeches by 190 leaders from a total of 58 countries. The original speeches are in 35 different languages and have been translated into English.\footnote{Translation via Google Translate.} They range from 139 to 1,468 words long, with a mean length of 414.75 and a median length of 410 words. The GPD coders generated different variables from the speech data, and I will use one of these for analysis: the type of speech (\textit{speechtype}). The measure has four categories: \textit{international}, where the audience is foreign and the speech is given outside the speaker's country of origin; \textit{campaign}, which refer to speeches given on the campaign trail; \textit{ribboncutting}, given to a local audience to introduce a new idea or project; and \textit{famous}, a notable and well-known speech that displays the president's leadership and speaking abilities.\footnote{The distribution of the categories is as follows: 220 speeches are `famous' (26.1\%), 218 are `international' (25.9\%), 213 are `ribboncutting' (25.3\%), and 192 are `campaign' (22.7\%).} From these four categories, I focus on \textit{international} speeches to illustrate the synthetic data approach. This is a good category to test the method with because it is highly contextual but there are also specific patterns in the text that identify it as international.

The challenges with this dataset are two. First, speeches are long (some have over 50,000 words), and Transformers-based models are limited to 512 tokens \citep{timoneda_jop}.\footnote{512 tokens, which are parts of words or \textit{subwords}, usually correspond to around 275-350 words.} I set a maximum length of 512 tokens for the model, and select the first 512 tokens from every speech for simplicity. The second problem is sample size. With only 843 speeches across four categories, this leaves only around 200 speeches per category to train \textit{and validate} the model.\footnote{Usually, in cross-validation, we set aside 20 percent of observations for validation and use the remaining 80 percent for training. The cross-validations used in this article use these shares.} This is just over the threshold of acceptability for models to have sufficient data to train on efficiently and effectively due to batching issues \citep{timoneda_jop}.  

\subsection*{Generating Data for the Synthetic Imputation Procedure}

To apply the synthetic imputation procedure, I draw random samples for each of the two datasets detailed above. First, for the political nostalgia data, I draw three samples from the full set of 151 nostalgic sentences. Each of these samples will be used to test performance with different numbers of true samples. The first sample contains 50 true nostalgic texts, the second 75, and the third 100. I use the first sample (50) to generate 101 synthetic texts using GPT-4o. With the second sample (75), I generate 76 synthetic texts. I then repeat the process to generate 51 synthetic texts with the third sample (100) --- all samples sum to 151, the number of original texts.\footnote{All samples include the same number of original texts to make sure we can make direct and valid comparisons with true model performance.} For international speeches, I also draw samples of 50, 75, 100, and add a sample of 150, and generate the corresponding 150, 125, 100 and 50 synthetic speeches. For both tasks, I will use the full original dataset to measure true model performance.

To generate the GPT-4o prompt, applied researchers should be very explicit with their instructions to generate text that is different both in terms of content and sentence structure from the examples provided to model in the prompt \citep{gao2023prompt, hatakeyama2023prompt, white2023prompt}. Specifically, I propose using a general `system' instruction and a set of \textit{five examples} drawn at random from the sample in each iteration. This approach reinforces the idea that few-shot learning tends to produce the best results with GPT-4o \citep[][(citation omitted)]{ornstein2022train, gilardi2023chatgpt}. The general instruction for the nostalgia application is: \textit{``Generate a nostalgic text in english based on the examples below. [new line] Make sure the new text is one whole paragraph and is different in content from the examples, but still has a nostalgic tone. The length of the new text should be similar to the examples. Names, countries and topics should also be different. Here are the five examples: [new line] Example 1: \{\} [new line] Example 2: \{\} [new line] Example 3: \{\} [new line] Example 4: \{\} [new line] Example 5: \{\}.''}\footnote{[new line] represents `\textbackslash n' in Python. Empty curly braces \{\} represent empty spaces where the examples are set every time in the loop. Two further notes about this prompt. First, 512 tokens (the maximum allowed in a RoBERTa model) correspond to around 275-350 words. I ask for a slightly longer speech to ensure that they are long enough for the model. Second, the example here allows for differences to be generated at the country level and via different names. Other applications will involve changes to different parts of the text. It is important that researchers identify parts of their text for substitution and provide explicit instructions to the model to do that.} The general instruction for the speeches application replicates this one.\footnote{The exact text reads: \textit{``Generate the first 500 words of a speech in English. [new line] Make sure the new text is one whole paragraph and is different in content from the examples, but reflects the international nature and tone of the speech. The length of the new text should be similar to the examples. Names, countries and topics should also be different. Here are the five examples: [new line] Example 1: \{\} [new line] Example 2: \{\} [new line] Example 3: \{\} [new line] Example 4: \{\} [new line] Example 5: \{\}''}.} The five examples are drawn at random \textit{each} time from the sub-sample of original speeches and passed inside the prompt.\footnote{To generate the output with GPT-4, I used the latest OpenAI API in Python, passing the prompt in italics above as the `system' prompt. This is the methodology recommended by OpenAI.} Drawing examples using this approach creates variation across new texts and minimizes overfitting. 

\begin{figure}[!b]
    \centering
    \includegraphics[width=17cm]{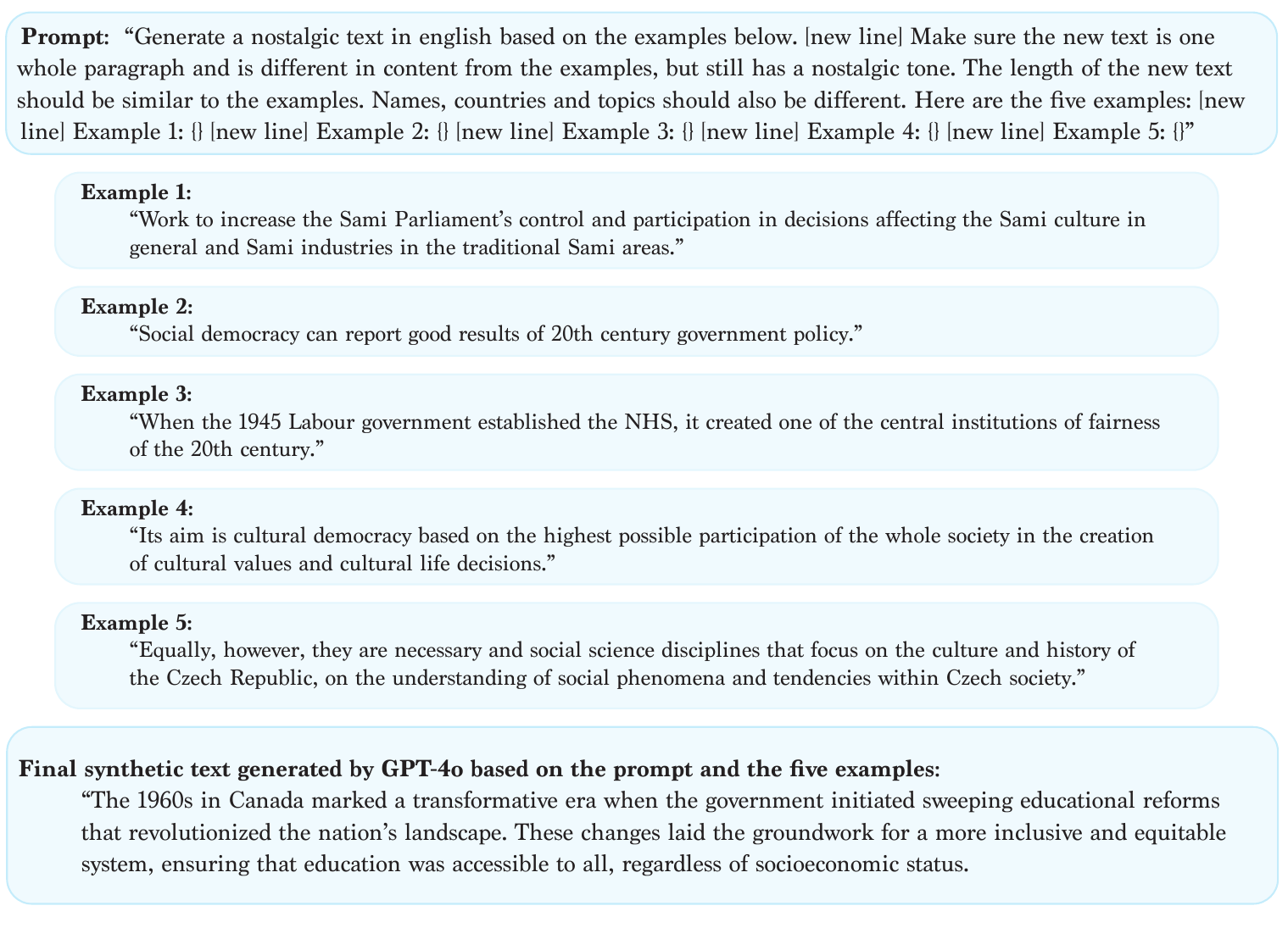}
    \caption{\small Flow chart for an example of synthetic text generation using the synthetic imputation approach (note: each example is included in the prompt in place of the `\{\}').}
    \label{fig:chart}
\end{figure}

Figure \ref{fig:chart} displays a flow chart describing the synthetic imputation approach from prompt creation to synthetic text output. The example provided is from the nostalgia dataset and represents one of the iterations in the synthetic imputation approach described above. Importantly, it shows how careful prompting and selecting five random examples produces a new synthetic text instance that retains substantive meaning across the examples but differs substantively both grammatically and semantically. This balance is key to prevent overfitting when finetuning the classifier.

\section*{Analysis and Results}

It is key that synthetic performance be neither below \textit{nor above} the performance of the model trained on the original dataset. If synthetic data performance is below true data performance, it means the new approach is not providing sufficient information to the model to identify the different categories correctly. Conversely, if synthetic performance is \textit{higher} than true performance, it means that the synthetic tests are too similar to the sample of true texts, leading to \textit{overfitting} during training. The tests below identify (1) the amount of true instances required for comparable performance between the synthetic approach and an original dataset, and (2) the amount of overfitting when true instances are too few. They also provide a comparison to the SSMBA data augmentation approach.

I finetune a set of RoBERTa models for each application. First, for the political nostalgia task, I finetune a total of ten different RoBERTa-large classifiers: three for the synthetic imputation approach, three for the baselines without the synthetic data, three for the SSMBA alternative approach, and the true model with all original data.\footnote{Following \citet{timoneda_jop}, I use the English-based `Roberta-large' model with a batch size of 16, the recommended learning rate of $3e-5$ for political nostalgia and $5e-6$ for speeches, 6 epochs with an early stopping mechanism to prevent overfitting, 10 warm-up steps, and a weighted ADAM optimizer with an epsilon value of $1e-6$. Regarding the two different learning rates, the cross-lingual model requires values at or around $5e-6$, while English RoBERTa requires them at or around $3e-5$.} Each set of three models  cover the different data structures with 50, 100, and 150 original observations, but differs in terms of whether and which type of synthetic data are included. The synthetic imputation models include synthetic text data generated with the approach proposed in this article.\footnote{The exact structures are (1) 50 original / 101 synthetic; (2) 75 original / 76 synthetic; (3) 100 original / 51 synthetic.} The SSMBA models include the same amount of synthetic data as the synthetic imputation models, but with text generated using the SSMBA approach.\footnote{Again, for clarity, the exact structures for the SSMBA models are (1) 50 original / 101 synthetic; (2) 75 original / 76 synthetic; (3) 100 original / 51 synthetic. The synthetic data in this case is generated using the SSMBA code and approach.} The baseline models, on the other hand, include only the original sets of 50, 75, and 100 observations and no synthetic data. The reason for including the three baseline models is to reflect model performance in cases where researchers have reduced or insufficient amounts of data and the trade-offs involved when not using synthetic imputation. They also highlight the importance of including synthetic data to improve model performance when original data is scarce, as not doing so severely affects model performance. Lastly, I run the model on the full sample, which includes all 151 nostalgic texts, to estimate true performance on the full dataset.

\begin{figure}[!t]
    \centering
    \includegraphics[width=14cm]{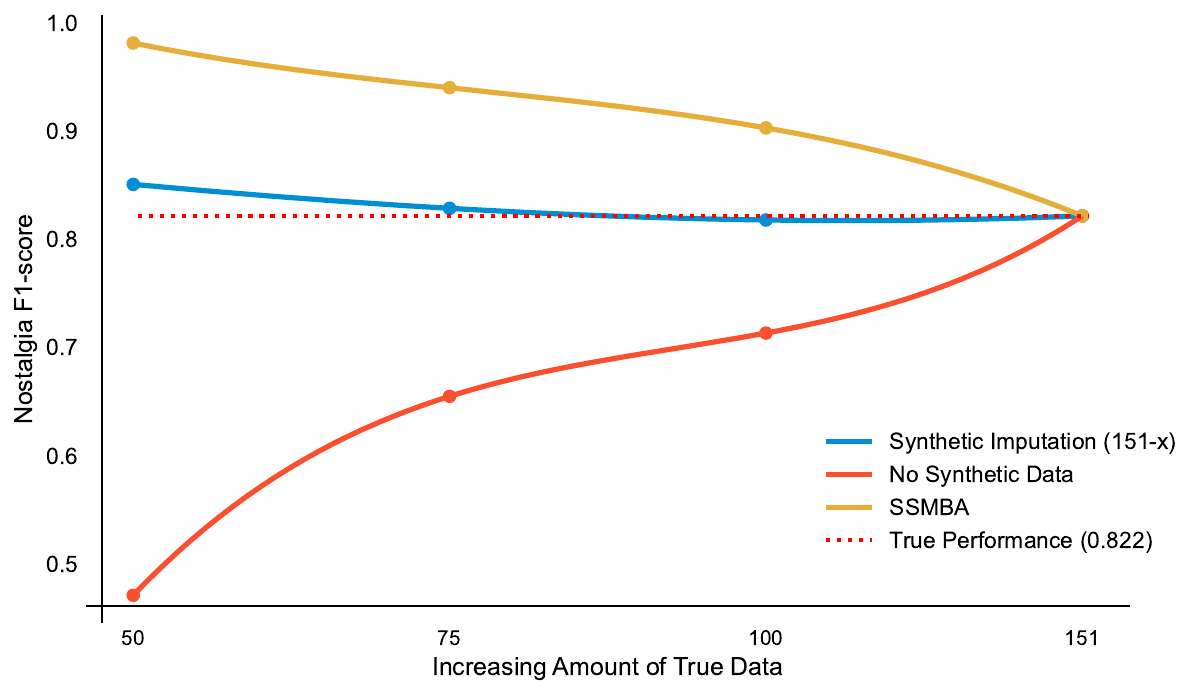}
    \caption{\small Results for the Nostalgia category from the Political Nostalgia dataset. Each $F1-$score average is derived from 10-time repeated 10-fold CV runs for synthetic, baseline and true models.}
    \label{fig:nost}
\end{figure}

\begin{figure}[!b]
    \centering
    \includegraphics[width=14cm]{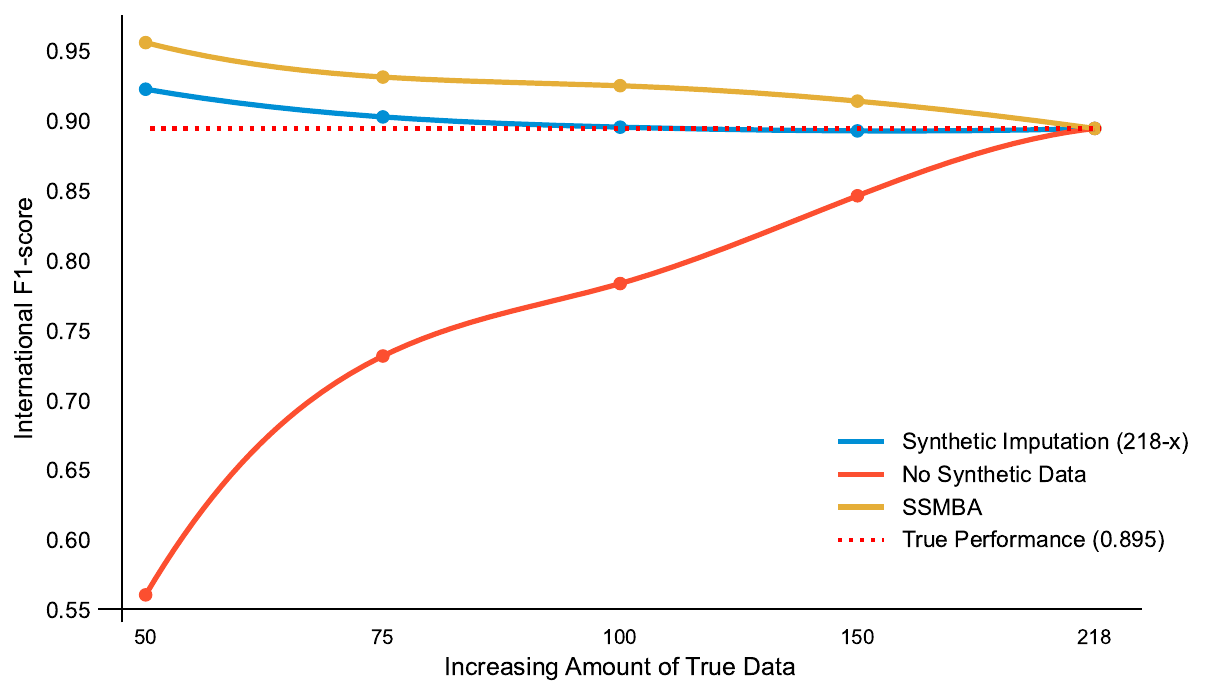}
    \caption{\small Results for the International Speeches category from the Global Populism Dataset. Each $F1-$score average is derived from 10-time repeated 10-fold CV runs for synthetic, baseline and true models.}
    \label{fig:speeches}
\end{figure}

For the second application, international speeches, I follow the same approach and run four models using the synthetic imputation approach,\footnote{(1) 50 original / 168 synthetic; (2) 75 original / 143 synthetic; (3) 100 original / 118 synthetic; (4) 150 original / 68 synthetic.}, four models using SSMBA data augmentation, and four baseline models with only 50, 75, 100 and 150 original observations but no synthetic data. I also run the original model with the full dataset (218 international speeches) to serve as the true performance counterfactual. For each model in both tasks, I run 10-times repeated 10-fold cross-validation (CV) to ensure more precise performance estimates over multiple CV runs \citep{timoneda_jop}.\footnote{This method effectively performs 10-fold CV ten different times for a total of 100 models, and its results are more precise that one-time CV.} In Figures \ref{fig:nost} and \ref{fig:speeches}, I report mean $F1$-scores for these sets of models for both the `nostalgic' and `international speeches' datasets, respectively. 

Figure \ref{fig:nost} reports the results with the political nostalgia dataset. $F1$ performance is close to true model performance (dashed red line) for all synthetic imputation models (blue line), which also significantly outperform both the SSMBA models and the models without synthetic data. True model performance with all training data available stands at 0.822 for the nostalgia category (texts coded as `nostalgic').\footnote{This is reflected both in the horizontal dashed red line and the performance scores at 151 observations (x-axis).} With only 50 original observations and 151 synthetic texts, the RoBERTa classifier shows very moderate levels of overfitting, with an $F1$-score of 0.851, or 3.5\% higher than true performance. Compare this with the 50-text model \textit{without} synthetic imputation, which produces a low $F1$-score of 0.471 for the nostalgia category. This represents a 74.5\% improvement when using the synthetic imputation approach with a true sample of 50 observations.\footnote{The gain is calculated from 0.822, not 0.851, to penalize for slight overfitting.} Moreover, synthetic imputation also reduces overfitting significantly when compared to SSMBA, whose texts are too similar to each other and thus inflate model performance on the training data, compromising out-of-sample performance. SSMBA with only 50 original observations and 168 synthetic ones produces an $F1$-score of 0.982. Synthetic imputation thus reduces overfitting by 13.3\% with smaller samples and, importantly, produces performance levels much closer to true model performance. 

These category-specific differences also affect overall model performance significantly. The overall $F1$-score for the original model is 0.879 (the `not nostalgic' category has higher $F1$-scores than the `nostalgic' category). With 50 original texts plus synthetic imputation, overall model performance stands at 0.901, which means the model as a whole overfits the data only slightly---by 2.5\%. Conversely, without synthetic imputation, the overall $F1$-score is 0.711, or a 19.1\% decrease. Moderate levels of overfitting, therefore, may be acceptable in cases where there are few original texts and more synthetic ones, for two reasons. First, model performance is very close to true performance, much closer than not using synthetic imputation. Second, the amount of overfitting is always in one direction (overestimating performance) and consistently around 2-4\%, so researchers can report a penalized level of performance.\footnote{This will not aid the small bias induced when generalizing to unseen, out-of-sample data, but considering the level of overfitting is small, the measurement error will also be small. The extent to which the error may affect results in small original samples with more synthetic observations requires further research.}

More importantly, Figure \ref{fig:nost} confirms the finding that performance is equivalent between the original sample and the sample with only 75 original texts plus synthetic imputation. The $F1$-score for this model is 0.829, which is only 0.007 points higher than the original model performance (0.822), a difference of less than one percent. Considering the standard deviations for both scores are 0.01, both scores are within one standard deviation of each other and thus statistically indistinguishable (overall model performance is also equivalent, with $F1$-scores of 0.879 and 0.887, respectively). Contrast this with the baseline with no synthetic data, whose $F1$-score is only 0.655. Synthetic imputation improves performance by 25.5\% in this data structure while producing an $F1$-score that mirrors true model performance. SSMBA continues to overfit the data, with an $F1$-score of 0.940. The same applies to the dataset with 100 original observations and 51 synthetic ones, whose performance is also indistinguishable from true performance with an $F1$-score of 0.818. Overall model performance is exactly the same for both the 100 text model and the original model, 0.879. Therefore, this application suggests that at and over 75 true observations plus synthetic imputation, model performance is equivalent to the original model with the full set of training data. It also shows that the synthetic imputation approach produces significant performance gains over the other two alternatives, SSMBA and including no synthetic data in model finetuning.

Figure \ref{fig:speeches} reports the $F1$-scores for all the international speeches models. It again provides strong evidence that the synthetic imputation approach can provide a solution in cases with insufficient training data. True performance is 0.895 ($F1$-score) when all 218 international speech observations are included in the training data. In the Figure, this is reflected with the dashed red line and the points in the plot furthest to the right (x=218). First, baseline model performance decreases progressively as data become sparser in the training data (red line). With 150 speeches, the $F1$-score goes down to 0.847; with 100 speeches, to 0.784; with 75 speeches, to 0.732; and with only 50 speeches, to 0.561. These results again show the high performance trade-offs when data are scarce. When adding synthetic data through synthetic imputation (blue line), performance improves significantly. With 150 original speeches and 50 synthetic speeches, the international $F1$-score is 0.893, well within one standard deviation for the CV runs (0.03). With 100 synthetic and original speeches, the $F1$-score is 0.895, exactly the same as true model. With 75 original and 125 synthetic speeches, the international $F1$-score increases to 0.902, which is within a one standard deviation of the true model's $F1$-score (sd=0.007). Lastly, with only 50 original speeches and 150 synthetic ones, performance does increase to an $F1$-score of 0.923, which is 3.1 percent higher than the true model and again indicates very moderate overfitting. This application confirms, however, that overfitting is both relatively small, predictable, and easy to penalize, and the benefits of it outweigh the performance costs of not using synthetic imputation with only 50 original texts. Lastly, note again that synthetic imputation improves on SSMBA performance significantly across all synthetic models ---synthetic imputation reduces overfitting by 6\%, on average. As with the nostalgia dataset, SSMBA tends to produce highly similar texts that lead to higher levels of overfitting on the training data. In this case, the overall difference between SSMBA and synthetic imputation is smaller, but this is in large part due to ceiling effects: true model performance is higher (0.892), so synthetic imputation has less overfitting to correct. Therefore, \textit{both} applications confirm two relevant findings. First, and most important: at and over 75 true observations plus synthetic imputation, model performance is equivalent to the original model with the full set of training data. Second, even with only 50 original observations, a very moderate level of overfitting may be acceptable considering the performance benefits for the overall model.

How generalizable are these results to other classification tasks? I argue that they generalize well for various reasons. First, both sets of Transformers models used here (RoBERTa and GPT-4) are deep learning neural networks that find patterns in the data to produce a classification (RoBERTa) or generate text (GPT-4). There are no explicit rules used in the classification exercise. This means that both models should adapt easily to other types of text, especially considering their state-of-the-art performance levels with highly nuanced and context-specific text \citep{timoneda_jop}. Performance will vary with the difficulty of the task, but that will apply to all models, including the true model. Second is the simplicity of the prompt. If the researcher provides a general prompt instructing the model to vary the text according to modifiable parameters (names, countries, laws, cities, etc.), and a random sample of five examples from the original text for every newly generated instance, we can expect synthetic performance to resemble the results outlined in this article.

\section*{Conclusion}

The main takeaway from this article is that synthetic data produces no statistically different performance than true data when the number of original texts is at least 75. Below 75, the model increasingly overfits the data, although at a relatively modest rate. In fact, overfitting at 50 original observations is preferable to the drop-off in performance without synthetic data. The predictable direction and magnitude of the overfitting at 50 original observations allows researchers to report penalized performance scores and be aware of the measurement bias introduced in downstream tasks. Researchers should weigh these trade-offs when faced with their own fine-tuning task and account for potential biases in downstream analyses arising from known overfitting. Above 75 original texts, however, researchers can confidently use the synthetic imputation approach introduced in this article with limited overfitting concerns. Doing so will produce models that generate much more accurate annotated data for downstream tasks. This is especially important for two types of projects: first, those that otherwise would be abandoned for lack of data; and second, projects where some underrepresented categories must be merged together into a broader category, even if the broader category is much less precise --thus inducing measurement bias anyway. By applying the synthetic imputation approach, many of these projects now can improve classifier performance without inducing overfitting, thus minimizing overall measurement bias from low amounts of data in specific categories.

\vspace{1cm}

\setstretch{1}
\bibliography{synthetic}

@article{timoneda_jop,
	title = {BERT, RoBERTa, or DeBERTa?},
	journal = {Journal of Politics},
	author = {Timoneda, Joan C. and  Vallejo Vera, Sebastián},
	year = {2025},
}

@inproceedings{hawkins2019measuring,
  title={Measuring populist discourse: The global populism database},
  author={Hawkins, Kirk A and Aguilar, Rosario and Silva, Bruno Castanho and Jenne, Erin K and Kocijan, Bojana and Kaltwasser, Crist{\'o}bal Rovira},
  booktitle={EPSA Annual Conference in Belfast, UK, June},
  pages={20--22},
  year={2019}
}

@techreport{ornstein2022train,
  title={How to train your stochastic parrot: Large language models for political texts},
  author={Ornstein, Joseph T and Blasingame, Elise N and Truscott, Jake S},
  year={2022},
  institution={Working Paper}
}

@article{liu2019roberta,
  title={Roberta: A robustly optimized bert pretraining approach},
  author={Liu, Yinhan and Ott, Myle and Goyal, Naman and Du, Jingfei and Joshi, Mandar and Chen, Danqi and Levy, Omer and Lewis, Mike and Zettlemoyer, Luke and Stoyanov, Veselin},
  journal={arXiv preprint arXiv:1907.11692},
  year={2019}
}

@inproceedings{zhou2021isobn,
  title={Isobn: Fine-tuning bert with isotropic batch normalization},
  author={Zhou, Wenxuan and Lin, Bill Yuchen and Ren, Xiang},
  booktitle={Proceedings of the AAAI Conference on Artificial Intelligence},
  volume={35},
  number={16},
  pages={14621--14629},
  year={2021}
}

@article{wei2021text,
  title={Text augmentation in a multi-task view},
  author={Wei, Jason and Huang, Chengyu and Xu, Shiqi and Vosoughi, Soroush},
  journal={arXiv preprint arXiv:2101.05469},
  year={2021}
}

@article{wei2019eda,
  title={Eda: Easy data augmentation techniques for boosting performance on text classification tasks},
  author={Wei, Jason and Zou, Kai},
  journal={arXiv preprint arXiv:1901.11196},
  year={2019}
}

@inproceedings{guo2020nonlinear,
  title={Nonlinear mixup: Out-of-manifold data augmentation for text classification},
  author={Guo, Hongyu},
  booktitle={Proceedings of the AAAI Conference on Artificial Intelligence},
  volume={34},
  number={04},
  pages={4044--4051},
  year={2020}
}

@article{nie2020named,
  title={Named entity recognition for social media texts with semantic augmentation},
  author={Nie, Yuyang and Tian, Yuanhe and Wan, Xiang and Song, Yan and Dai, Bo},
  journal={arXiv preprint arXiv:2010.15458},
  year={2020}
}

@article{ng2020ssmba,
  title={SSMBA: Self-supervised manifold based data augmentation for improving out-of-domain robustness},
  author={Ng, Nathan and Cho, Kyunghyun and Ghassemi, Marzyeh},
  journal={arXiv preprint arXiv:2009.10195},
  year={2020}
}

@article{devlin2018bert,
  title={Bert: Pre-training of bidirectional transformers for language understanding},
  author={Devlin, Jacob and Chang, Ming-Wei and Lee, Kenton and Toutanova, Kristina},
  journal={arXiv preprint},
  year={2018}
}

@article{hobbs2019effects,
  title={Effects of divisive political campaigns on the day-to-day segregation of Arab and Muslim Americans},
  author={Hobbs, William and Lajevardi, Nazita},
  journal={American Political Science Review},
  volume={113},
  number={1},
  pages={270--6},
  year={2019},
  publisher={Cambridge University Press}
}

@article{roberts2016navigating,
  title={Navigating the local modes of big data: the case of topic models},
  author={Roberts, Margaret E and Stewart, Brandon M and Tingley, Dustin},
  journal={Soc. Sci},
  volume={4},
  year={2016}
}

@article{egami2023using,
  title={Using Large Language Model Annotations for Valid Downstream Statistical Inference in Social Science: Design-Based Semi-Supervised Learning},
  author={Egami, Naoki and Jacobs-Harukawa, Musashi and Stewart, Brandon M and Wei, Hanying},
  journal={arXiv preprint},
  year={2023}
}

@article{rodriguez2023embedding,
  title={Embedding regression: Models for context-specific description and inference},
  author={Rodriguez, Pedro L and Spirling, Arthur and Stewart, Brandon M},
  journal={American Political Science Review},
  pages={1--20},
  year={2023},
  publisher={Cambridge University Press}
}

@article{rodriguez2022word,
  title={Word embeddings: What works, what doesn’t, and how to tell the difference for applied research},
  author={Rodriguez, Pedro L and Spirling, Arthur},
  journal={The Journal of Politics},
  volume={84},
  number={1},
  pages={101--115},
  year={2022},
  publisher={The University of Chicago Press Chicago, IL}
}

@article{grimmer2021machine,
  title={Machine Learning for Social Science: An Agnostic Approach},
  author={Grimmer, Justin and Roberts, Margaret E and Stewart, Brandon M},
  journal={Annual Review of Political Science},
  volume={24},
  pages={395--419},
  year={2021},
  publisher={Annual Reviews}
}

@article{chang2020using,
  title={Using word order in political text classification with long short-term memory models},
  author={Chang, Charles and Masterson, Michael},
  journal={Political Analysis},
  volume={28},
  number={3},
  pages={395--411},
  year={2020},
  publisher={Cambridge University Press}
}

@article{grimmer_stewart_2013, 
title={Text as Data: The Promise and Pitfalls of Automatic Content Analysis Methods for Political Texts}, 
volume={21}, 
DOI={10.1093/pan/mps028}, number={3}, journal={Political Analysis}, publisher={Cambridge University Press}, author={Grimmer, Justin and Stewart, Brandon M.}, year={2013}, pages={267–297}}

@article{barbera2021automated,
  title={Automated text classification of news articles},
  author={Barber{\'a}, Pablo and Boydstun, Amber E and Linn, Suzanna and McMahon, Ryan and Nagler, Jonathan},
  journal={Political Analysis},
  volume={29},
  number={1},
  pages={19--42},
  year={2021},
  publisher={Cambridge University Press}
}

@article{catalinac2016pork,
  title={From pork to policy: The rise of programmatic campaigning in Japanese elections},
  author={Catalinac, Amy},
  journal={The Journal of Politics},
  volume={78},
  number={1},
  pages={1--18},
  year={2016},
  publisher={University of Chicago Press Chicago, IL}
}

@article{timoneda2025behind,
  title={Behind the mask: Random and selective masking in transformer models applied to specialized social science texts},
  author={Timoneda, Joan C. and Vallejo Vera, Sebastián},
  journal={PloS one},
  volume={20},
  number={2},
  pages={e0318421},
  year={2025},
  publisher={Public Library of Science San Francisco, CA USA}
}

@article{wettig2022should,
  title={Should you mask 15\% in masked language modeling?},
  author={Wettig, Alexander and Gao, Tianyu and Zhong, Zexuan and Chen, Danqi},
  journal={arXiv preprint arXiv:2202.08005},
  year={2022}
}

@article{muller2024nostalgia,
  title={Nostalgia in European party politics: a text-based measurement approach},
  author={M{\"u}ller, Stefan and Proksch, Sven-Oliver},
  journal={British Journal of Political Science},
  volume={54},
  number={3},
  pages={993--1005},
  year={2024},
  publisher={Cambridge University Press}
}

@article{gilardi2023chatgpt,
  title={ChatGPT outperforms crowd workers for text-annotation tasks},
  author={Gilardi, Fabrizio and Alizadeh, Meysam and Kubli, Ma{\"e}l},
  journal={Proceedings of the National Academy of Sciences},
  volume={120},
  number={30},
  pages={e2305016120},
  year={2023},
  publisher={National Academy Sciences}
}

@article{gao2023prompt,
  title={Prompt engineering for large language models},
  author={Gao, Andrew},
  journal={SSRN 4504303},
  year={2023}
}

@article{white2023prompt,
  title={A prompt pattern catalog to enhance prompt engineering with chatgpt},
  author={White, Jules and Fu, Quchen and Hays, Sam and Sandborn, Michael and Olea, Carlos and Gilbert, Henry and Elnashar, Ashraf and Spencer-Smith, Jesse and Schmidt, Douglas C},
  journal={arXiv preprint arXiv:2302.11382},
  year={2023}
}

@article{hatakeyama2023prompt,
  title={Prompt engineering of GPT-4 for chemical research: what can/cannot be done?},
  author={Hatakeyama-Sato, Kan and Yamane, Naoki and Igarashi, Yasuhiko and Nabae, Yuta and Hayakawa, Teruaki},
  journal={Science and Technology of Advanced Materials: Methods},
  volume={3},
  number={1},
  pages={2260300},
  year={2023},
  publisher={Taylor \& Francis}
}

@article{rice2021corpus,
  title={Corpus-based dictionaries for sentiment analysis of specialized vocabularies},
  author={Rice, Douglas R and Zorn, Christopher},
  journal={Political Science Research and Methods},
  volume={9},
  number={1},
  pages={20--35},
  year={2021},
  publisher={Cambridge University Press}
}
\bibliographystyle{apsr}
\nocite*{}

\end{document}